\documentclass[letterpaper, 10 pt, conference]{ieeeconf}

\usepackage{mathtools}
\usepackage[pdftex,final]{graphicx}
\usepackage[english]{babel}
\usepackage[latin1]{inputenc}
\usepackage{balance}

\usepackage[bookmarks=true]{hyperref}
\usepackage{url}
\usepackage{amsmath}
\usepackage{xcolor}
\usepackage{flushend}
\usepackage{empheq}

\usepackage{caption}
\usepackage{subfigure}

\usepackage{algorithm}
\usepackage[noend]{algpseudocode}

\usepackage{textpos}

\newcommand{\pre}[1]{{}^{#1}\!}
\newcommand{\sam}[1]{\pre{k}#1}
\newcommand{\giv}{\,|\,} 

\usepackage{setspace}
\let\Algorithm\algorithm
\renewcommand\algorithm[1][]{\Algorithm[#1]\setstretch{1.4}}

\makeatletter
\def\BState{\State\hskip-\ALG@thistlm}
\makeatother

\definecolor{darkblueish}{RGB}{64, 87, 116}
\definecolor{blueish}{RGB}{103, 135, 176}
\definecolor{greenish}{RGB}{177, 177, 123}
\definecolor{reddish}{RGB}{ 205, 102, 7}
\definecolor{orangeish}{RGB}{246, 160, 61}
\definecolor{somethingilike}{RGB}{250, 232, 184}
\colorlet{bluegreenish}{blueish!50!greenish}

\hypersetup{
    colorlinks,
    linkcolor={reddish!70!black},
    citecolor={blueish!70!black},
    urlcolor={blueish!70!black}
}
\usepackage[all]{hypcap}

\usepackage{todonotes}

\makeatletter
\def\input@path{{../figures/}}
\makeatother
\graphicspath{{../figures/}}

\makeatletter
\let\NAT@parse\undefined
\makeatother
\usepackage[numbers]{natbib}
                                                                                       
\IEEEoverridecommandlockouts                             
\title{\LARGE \bf
Depth-Based Object Tracking Using a Robust Gaussian Filter
}

\author{Jan Issac$^{1*}$, Manuel W\"uthrich$^{1*}$, Cristina Garcia Cifuentes$^{1}$,\\ Jeannette Bohg$^{1}$, Sebastian Trimpe$^{1}$ and Stefan Schaal$^{1,2}$ 
  \thanks{$^{1}$ Autonomous Motion Department at the Max-Planck-Institute for Intelligent Systems, T\"ubingen, Germany
    Email: {\tt\small   first.lastname@tuebingen.mpg.de}}%
  \thanks{$^{3}$ Computational Learning and Motor Control lab at the
  University of Southern California, Los Angeles, CA, USA}%
}

\begin{document}

\maketitle
\thispagestyle{empty}
\pagestyle{empty}

\begin{abstract}
 We consider the problem of model-based 3D-tracking of objects given dense
depth images as input. Two difficulties preclude the application of a 
standard Gaussian filter to this problem. First of all, depth
sensors are characterized by fat-tailed measurement noise. To address
this issue, we show how a recently published robustification method 
for Gaussian filters can be applied to the problem at hand.
Thereby, we avoid using heuristic outlier detection methods
that simply reject measurements if they do not match the 
model.
%
%
Secondly, the computational cost of the standard Gaussian filter is
prohibitive due to the high-dimensional measurement, i.e. the depth image.
To address this problem, we propose an approximation to reduce the
computational complexity of the filter.
In quantitative experiments on real data 
we show how our method clearly
outperforms the standard Gaussian filter. Furthermore, we compare its performance 
to a particle-filter-based tracking method, 
and observe comparable computational efficiency and improved accuracy and smoothness of the estimates.
\end{abstract}


%

\begin{textblock*}{100mm}(.\textwidth,-11.3cm)
 \begin{spacing}{.8}
 {\fontsize{8pt}{2pt}\selectfont \sffamily
\noindent 2016 IEEE International Conference on\\
Robotics and Automation (ICRA)\\
Stockholm, Sweden, May 16-21, 2016}
\end{spacing}
\end{textblock*}%

\section{Introduction}
As a robot interacts with its surroundings, it needs to  continuously estimate 
its own and the environment's state. Only then, this estimate can
provide timely feedback to low-level controllers or motion planners to choose the
appropriate next movement. More concretely, for purposeful and robust
manipulation of an object, it is  crucial for the robot to know the
location of its own manipulator and the target object. This
information is not directly observable but has to be inferred from
noisy sensory data. 
%
%
%

In this paper, we address the problem of continuously inferring the
6-degree-of-freedom pose and velocity of an object from dense depth images.
A 3D mesh model of the considered object is assumed to be known.
Sensors yielding depth images, such as the Microsoft Kinect or the Asus Xtion,
are widespread within robotics and provide a great amount of information about
the state of the environment. However, working with such sensors can be challenging
due to their noise characteristics (e.g. presence of outliers) and the high 
dimensionality of the measurement. These difficulties have thus far precluded 
the direct application of Gaussian filtering methods to image measurements.

\subsection{Related Work}


For low-level sensors such as
joint encoders, inertial measurement units and torque sensors, estimation 
is typically performed using Gaussian filters (GF) \cite{gf, sarkka}.
The most well known members of the family of GFs are the Extended Kalman Filter 
(EKF) \cite{ekf} and the Unscented Kalman Filter (UKF) \cite{ukf}.

While GFs are the preferred estimation method for low-level sensors, they
have rarely been applied directly to image data.
%
Instead, a large variety of different approaches has been applied to
model-based 3D object tracking. A comprehensive overview is given
by~\citet{Lepetit:2005}. Some notable recent methods that rely on matching
distinct features of the object model with features in the RGB measurement
have been proposed in \citep{hinterstoisser2012accv, choi}.  With the availability of cheap RGB-D
sensors such as the Microsoft Kinect or the Asus Xtion, the interest in the
problem of object tracking from depth data has increased greatly. 
A number of methods minimize some
energy function that captures the discrepancy between the current measurement
and state estimate \citep{ren, in_hand, dart,cvpr2014kyriazis}.
\citet{pauwels_imprecise_2014} combine energy minimization given
different visual cues with global feature-based re-initialization of the
tracker. 

\citet{hebert} provide an example of a GF-based method towards
object and manipulator tracking. However, this method does not process 
the depth data directly. Instead, features are extracted from 
the depth images and then passed to the GF as measurements.

%

 
Among filtering methods, particle filters (PF) \cite{gordon}
are used much more often for state estimation from cameras. 
The PF is well suited for two reasons. First of all,
it is non-parametric which makes it possible to model the
fat-tailed noise which typically corrupts image measurements, 
see for instance \cite{thrun_monte_carlo,fallon,pot}. Secondly, the heavy
computational requirements can be offset by modeling each pixel as an 
independent sensor \cite{pot} and by parallelization in the particles
\cite{fallon, changhyun}.

%
%
%
%
%

\subsection{Contributions}
%
%

These two properties of PFs do not hold for GFs. 
The standard GF fails completely for fat-tailed-distributed
measurements \cite{rgf}.  This is a serious problem since outliers are very
common in image data due to occlusions and other effects which
are difficult to model. To address this problem, we use a
robustification method for the GF which has recently been proposed in \cite{rgf}. The
resulting method is far more robust to occlusions and other unmodeled effects than a
standard GF.

Furthermore, naive application of the GF is very inefficient for
high-dimensional measurements, and parallelization is not as trivial as for the
PF. We use an observation model which factorizes in the pixels, similar to
\cite{fallon, pot}. The proposed method exploits this structure to reduce the
computational complexity and to allow for parallelization.

These results extend the domain of application of the GF to problems which have
previously been outside of its range. In many applications, it is preferable
to use a GF instead of a PF, since the latter suffers from particle deprivation
for all but very low dimensional state spaces. This issue manifests itself
as jumps in the estimates computed by the PF. This is especially problematic 
when the estimate is fed into a controller, e.g. for visual servoing.

We apply the proposed method to tracking the 6-degree-of-freedom pose and velocity 
of an object using only depth measurements from a consumer depth
camera. The proposed algorithm runs at the frame rate of the camera (30 Hz) on
just one CPU core. A quantitative comparison between the proposed method and a PF-based method \cite{pot}
is provided in the experimental section. We show that the proposed GF-based
method yields smoother estimates.

%

The quantitative experimental analysis is based on an object tracking dataset which we 
recorded for this purpose.
It consists of depth images recorded with an Asus Xtion camera, showing 
a person manipulating a set of rigid objects with different levels of
occlusions and motion velocities. This dataset is annotated with the
frame-by-frame ground-truth object pose as recorded with a Vicon motion capture
system. This dataset as well as the implementation of the proposed method are 
publicly available \cite{bot}.

\section{Gaussian Filtering}
In this section we briefly review filtering in general, and the GF in
particular. This will serve as a basis for the derivation of the proposed
algorithm in the following sections.

\subsection{Filtering}

Filtering is concerned with estimating the current state $x_t$ given all past 
measurements $y_{1:t}$. The posterior distribution of the current state 
$p(x_t \giv y_{1:t})$ can be computed recursively from the distribution of 
the previous state $p(x_{t-1} \giv y_{1:t-1})$. This recursion can be written 
in two steps, a prediction step
\begin{align}
 p(x_t \giv y_{1:t-1}) = 
   \int\limits_{x_{t-1}} p(x_t \giv x_{t-1}) p(x_{t-1} \giv y_{1:t-1}), \label{eq:prediction}
\end{align}
and an update step
\begin{align}
 p(x_t \giv y_{1:t}) = 
   \frac{p(y_t \giv x_t) p(x_t \giv y_{1:t-1})}
        {\int\limits_{x_t} p(y_t \giv x_t) p(x_t \giv y_{1:t-1})}. \label{eq:update}
\end{align}

These equations can generally not be solved in closed form \cite{earlyKushner}. 
The most notable exception is the Kalman Filter (KF) \cite{kalman1960new}, 
which provides the exact solution for linear Gaussian systems. Significant 
research effort has been invested into generalizing the KF to nonlinear 
dynamical systems. 

\subsection{Approximate Prediction Step}

Since the prediction \eqref{eq:prediction} does not admit a closed form 
solution for most nonlinear systems, the GF approximates the exact belief 
$p(x_t \giv y_{1:t-1})$ with a Gaussian distribution
$\mathcal{N}(x_t \giv \mu_{x_t}, \Sigma_{x_tx_t}$). The parameters are obtained 
by moment matching 
\begin{align}
 \mu_{x_t} &= \int\limits_{x_t} x_t p(x_t \giv y_{1:t-1}) \label{eq:mux} \\
 \Sigma_{x_tx_t} &= \int\limits_{x_t} (x_t - \mu_{x_t})(x_t - \mu_{x_t})^T p(x_t \giv y_{1:t-1}). \label{eq:sigmax}
\end{align}
When there is no analytic solution to these equations, they can be approximated
using numeric integration methods. Such methods are efficient in this setting
since samples from $p(x_t \giv y_{1:t-1})$ can be generated easily using
ancestral sampling: first we sample $\sam{x}_{t-1}$ from the previous belief
$p(x_{t-1} \giv y_{1:t-1})$, and then we sample $\sam{x}_t$ from the process
model $p(x_t \giv \sam{x}_{t-1})$.

Different numeric methods give rise to the different instances of the GF, such
as the EKF, the UKF and the Cubature Kalman Filter (CKF) \cite{ckf}. 
In all the experiments of this paper, we use the Unscented
Transform (UT) \cite{ukf} for numeric integration. However, for the ideas in
this paper, it is not relevant which particular integration method is used.

\subsection{Approximate Update Step}

In most practically relevant systems, the update step \eqref{eq:update} does not
admit an exact solution either. The approach taken here is different from the
prediction step, because it is rarely possible to sample from the posterior
\eqref{eq:update} efficiently \cite{new_perspective}. 

Therefore, the GF instead approximates the joint distribution 
$p(x_t, y_t \giv y_{1:t-1})$ 
with a Gaussian distribution \cite{new_perspective, sarkka}
\begin{align}
  q(x_t, y_t) = \mathcal{N}\left( 
    \begin{pmatrix} x_t \\ y_t \end{pmatrix} \Big|  
    \begin{pmatrix} \mu_{x_t} \\ \mu_{y_t} \end{pmatrix}, 
    \begin{pmatrix} \Sigma_{x_tx_t} & \Sigma_{x_ty_t} \\ \Sigma_{y_tx_t} & \Sigma_{y_ty_t} \end{pmatrix} 
    \right). \label{eq:joint_gaussian}
\end{align}
The desired approximation to the posterior distribution \eqref{eq:update} is
then obtained by conditioning on $y_t$, which is a simple operation for a
Gaussian \cite{barber}, 
\begin{align}
   q(x_t \giv y_t) = 
   \; \mathcal{N}(x_t \giv 
    &\mu_{x_t} + \Sigma_{x_ty_t} \Sigma_{y_ty_t}^{-1} (y_t - \mu_{y_t}), \nonumber  \\
    &~~~~~~~~\Sigma_{x_tx_t} - \Sigma_{x_ty_t} \Sigma_{y_ty_t}^{-1} \Sigma_{x_ty_t}^T). \label{eq:gaussian_filter_posterior}
\end{align}
The parameters are obtained by minimizing the KL divergence between the exact
and the approximate joint distributions \cite{new_perspective}
\begin{align}
  \mathrm{KL}[p(x_t, y_t \giv y_{1:t-1}) \giv q(x_t, y_t)].\label{eq:objective}
\end{align}

The optimal parameters $\mu_{x_t}$ and $\Sigma_{x_tx_t}$ are given by \eqref{eq:mux} and
\eqref{eq:sigmax} from the prediction step. The remaining optimal parameters are
\begin{align}
  \mu_{y_t} &= \int\limits_{y_t} y_t p(y_t \giv y_{1:t-1}) \label{eq:muy} \\ 
  \Sigma_{y_ty_t} &= \int\limits_{y_t} (y_t - \mu_{y_t})(y_t -\mu_{y_t})^T p(y_t \giv y_{1:t-1}) \label{eq:sigmay} \\ 
  \Sigma_{x_ty_t} &= \int\limits_{x_t, y_t} (x_t - \mu_{x_t})(y_t - \mu_{y_t})^T p(x_t, y_t \giv y_{1:t-1}).\label{eq:sigmaxy}
\end{align}
These integrals can be approximated efficiently, since samples from the joint
distribution can be generated easily. First we sample $\sam{x}_{t}$
from the prediction distribution $p(x_t|y_{1:t-1})$, and then we sample from the observation
model $p(y_t|\sam{x}_t)$.

\subsection{Computational Complexity of the Update Step}\label{sec:complexity}
For the tracking problem considered in this paper, the computational
bottleneck is the update step, where the high-dimensional measurement
is incorporated.

In the update step, we first approximate the integrals
\eqref{eq:muy}, \eqref{eq:sigmay} and \eqref{eq:sigmaxy} by sampling from 
the Gaussian prediction $p(x_t|y_{1:t-1})$, as explained in the previous section. 
To sample from a Gaussian
distribution requires computing the matrix square root of its covariance
matrix \eqref{eq:sigmax}, see \cite{ukf} for instance. The computational
cost of this operation is $O(N^3)$, where $N$ is the dimension of the state $x_t$ 
\cite{square_root_for_ukf}.

Once the integrals are approximated, the posterior \eqref{eq:gaussian_filter_posterior}
is obtained by 
conditioning on the measurement $y_t$. This operation requires the inversion
of the $M \times M$ matrix $\Sigma_{y_ty_t}$, where $M$ is the dimension of the
observation. This leads to a cubic complexity $O(M^3)$ in the dimension of the
observation as well \footnotemark.

%
%

\footnotetext{Some matrix computations can be carried out with a slightly lower
  computational complexity than the $O(N^3)$ required by the naive, element-wise
  approach. For instance, the Coppersmith-Winograd algorithm for $N \times N$
  matrix multiplication is $O(N^{2.376})$. In practice, however, such algorithms
  only represent an advantage for very large matrices, and they would affect
  both the standard GF and our method the same way. For simplicity, we hence
  compute all computational complexities using standard matrix operations.}
%

Hence, the computational complexity of the update step in the state dimension
$N$ and in the observation dimension $M$ is $O(N^3+M^3)$.

\section{Modeling}
\label{sec:model}
In this paper, 
the state to be estimated is the position $r_t$ and the orientation $o_t$ of the
tracked object. Since the orientation does not belong to a Euclidean space, we
choose to represent the state of the system by a differential change of
orientation $\delta o_t$ and position $\delta r_t$ with respect to a default
position $r_t^0$ and orientation $o_t^0$. This default pose is updated to
the mean of the latest belief after each filtering step. This way we 
make sure that the deviations $\delta o_t$ and $\delta r_t$ remain 
small and can therefore be treated as
members of a Euclidean space.

%
%

Additionally, we estimate the linear velocity $v_t$ and angular velocity
$\omega_t$ of the object.  The complete state is then
\begin{align}
  x = (\delta r_t, \delta o_t, v_t, \omega_t). \label{eq:state}
\end{align}

\begin{figure}[tb]
  \vspace{0.5cm}
  \centering
  \includegraphics[width=\columnwidth]{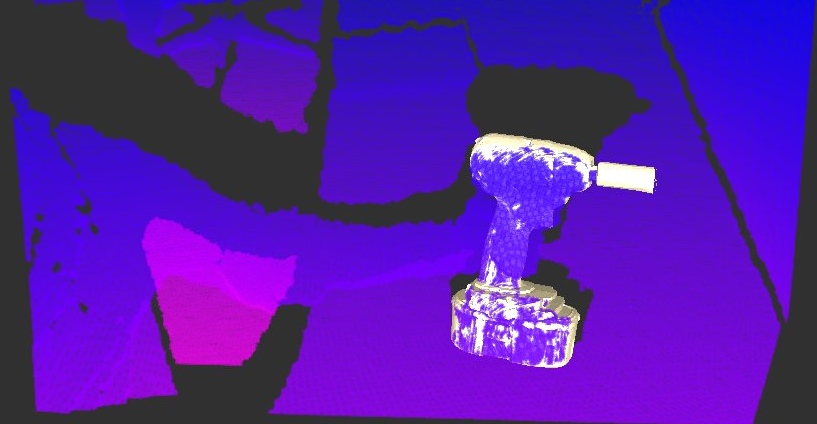}
  \caption{Depth-image visualized as a point cloud while tracking an impact 
           wrench.\label{fig:impact}}
\end{figure}

\subsection{Observation Model}
\label{sec:model:obs}

The sensor used is a depth camera. 
The observation is a range image $y_t$. A range image contains at each
pixel the distance from the camera to the closest object. It can be thought of
as an array of beams emitted by the camera, where at each beam the distance
traveled until the first intersection with an object is measured
(Figure~\ref{fig:impact}).

We treat each pixel $i$ as an independent sensor, the observation model
therefore factorizes
\begin{align}
  p(y_t \giv x_t) = \prod_i p(y_t^i \giv x_t).
\end{align}
The observation we would expect at a given pixel is the distance between the
camera and the tracked object along the corresponding beam. This distance
$d^i(x_t)$ can be computed easily for a given state $x_t$, since we have a mesh
model of the object. The observation could be modeled as the depth $d^i(x_t)$
corrupted with some Gaussian noise,
\begin{align}
 b(y_t^i \giv x_t) = \mathcal{N}(y_t^i \giv d^i(x_t), C) \label{eq:body_model}
\end{align}
where $C$ is the covariance matrix of the noise, which we assume to be equal
for all pixels.


However, the assumption of Gaussian noise is clearly violated in depth
images. The noise of the sensors is much more heavy tailed. Even more
importantly, there might be occlusions of the tracked object by other,
unmodeled objects.  The model above is not robust to such effects. Even a few
occluded sensors can introduce huge estimation errors, as we show in 
Section~\ref{sec:comparison_to_standard_gf}.

To address this problem, we introduce a second observation model, describing the
case where the measurement is produced by such unmodeled effects. To express
our ignorance about them, we choose the second observation model to be a uniform
distribution $t(\cdot)$ over the range of the sensor, which is roughly 0.5--7~m
for our depth sensor.
The complete observation model is a weighted sum of the body model $b(\cdot)$
and the tail model $t(\cdot)$ \begin{align}
  p(y_t^i \giv x_t) = (1-w) \, b(y_t^i \giv x_t) + w \, t(y_t^i) \label{eq:observation_model}
\end{align}
where $w$ is the probability of the measurement being produced by unmodeled
effects.

This simple extension is enough to account for outliers and occlusion, as shown
in the experimental section. Similar observation models have been used in
PF-based approaches to motion estimation from range data
\cite{thrun_monte_carlo,fallon,pot}. While standard PFs work with fat-tailed
observation models, this is not the case for the GF. In
Section~\ref{sec:fat_tails}, we apply a recent result \cite{rgf} to enable the
GF to work with this observation model.

\subsection{State Transition Model}

The state transition model is a simple linear model,
\begin{align}
 \delta r_t &= \delta r_{t-1} + v_{t-1}\\
 \delta o_t &= \delta o_{t-1} + \omega_{t-1}\\
 v_t &= v_{t-1} + \epsilon^v_t\\
 \omega_t &= \omega_{t-1} + \epsilon^\omega_t .
\end{align}
The velocity $v_t$ is perturbed by Gaussian noise $\epsilon^v_t$ at every time
step. It is then integrated into the position $\delta r_t$. Analogously, the
angular velocity $\omega_t$ is perturbed by noise $\epsilon^\omega_t$ and then
integrated into the orientation $o_t$.

\subsection{Difficulties of the Model}

There are two problems which preclude the application of a standard GF
to this model. First, the standard GF does not work with fat-tailed
observation models \cite{rgf}. Second, the standard GF is too computationally
expensive for this high-dimensional problem. In the following we address these
two issues.

\subsection{Notation}
In the remainder of the paper,
we only consider a single update step.
For ease of notation, we will not explicitly write the dependence
on all previous observations $y_{1 : t-1}$ anymore; it
is however implicitly present in all distributions. All
the remaining variables have the same time index $t$, 
which we can thus safely drop.
For example, $p(x_t ,y_t|y_{1 : t-1})$ becomes
$p(x,y)$ and $p(x_t |y_{1 : t})$ becomes $p(x|y)$, etc.

It is important to keep in mind that also the parameters 
computed in the following sections are time varying,
all computations are carried out at each time step.

\section{Gaussian Filtering with Fat Tails} 
\label{sec:fat_tails}

The GF depends on the observation model only through its
mean and covariance, it is incapable of capturing
more subtle features \cite{rgf}.
This means that the observation model \eqref{eq:observation_model} is
treated exactly as a Gaussian observation model with the same mean and
covariance. The covariance of  \eqref{eq:observation_model} is very large due to the uniform tail, so the
standard GF barely reacts to incoming measurements.

In \cite{rgf} we show that this problem can be addressed by replacing
the actual measurement $y$ by a virtual measurement
\begin{align}
\hspace{-0.18cm}
  \varphi(y,\mu_{y|b},\Sigma_{yy|b}) = \frac{
    \begin{pmatrix} 
      \mathcal{N}(y \giv \mu_{y|b}, \Sigma_{yy|b}) \\ 
      y \mathcal{N}(y \giv \mu_{y|b}, \Sigma_{yy|b}) \\ 
      t(y) 
    \end{pmatrix}}{
    (1\! -\! w) \, \mathcal{N}(y \giv \mu_{y|b}, \Sigma_{yy|b})
    + w \, t(y)}
  \label{eq:optimal_feature}
\end{align}
with the time-varying parameters
\begin{align}
  \mu_{y|b} &= \int_{x} \int_{y} 
    y b(y \giv x) p(x) \label{eq:muyb}\\ 
  \Sigma_{yy|b} &= \int_{x} \int_{y} 
    (y - \mu_{y|b})(y - \mu_{y|b})^T b(y \giv x) p(x). 
 \label{eq:sigmayb}
\end{align}
Since the different pixels are treated as independent sensors, this feature
is computed for each pixel independently.

The resulting algorithm, called the Robust Gaussian Filter (RGF) \cite{rgf},
corresponds to the standard GF where the observation models $p(y^i \giv x)$ are
replaced by virtual observation models $p(\varphi^i \giv x) = \int_{y^i}
p(\varphi^i \giv y^i) p(y^i \giv x)$, and the measurements from the sensor $y^i$
are replaced by the corresponding features $\varphi^i(y^i)$.  A step-by-step
description of this procedure is provided in Section \ref{sec:algorithm}.

\section{A Factorized Update}


As explained in Section~\ref{sec:complexity}, naive application of a GF to this
problem would lead to a computational complexity of $O(N^3+M^3)$, with $N$ being
the dimension of the state $x$ and $M$ the number of pixels of the depth
image $y$. Because $M$ is large, this complexity is prohibitive for real-time
applications.

Since the observation model \eqref{eq:observation_model} factorizes in the
$M$ measurements, the GF could be applied sequentially, incorporating the 
sensors one by one. We would carry out $M$ updates with a 
complexity of $O(N^3+1^3)=O(N^3)$ each, leading to 
a total complexity of $O(MN^3)$. Unfortunately, this implementation
can still be too slow for real-time usage. Furthermore, it cannot be 
parallelized because each update depends on the previous one.

In the following we derive an update for the GF with uncorrelated measurements
which has a complexity of $O(MN^2 + N^3)$. Furthermore, the computations are
independent for each sensor, making the algorithm eligible for parallelization.

\subsection{Explicit Approximation of the Observation Model}

The GF never explicitly computes  an approximate observation model $q(y \giv
x)$. Nevertheless, such a model is implied by the joint approximation $q(y, x)$.
For the subsequent derivation, it is necessary to make $q(y \giv x)$ explicit.
The objective function~\eqref{eq:objective} can be written as
%
%
%
\begin{align}
  & \mathrm{KL}[p(x, y) \giv q(x, y)] = 
    \int_{x, y} \log \left( \frac{p(x, y)}{q(x, y)}  \right) p(x, y) \notag \\
  & {} = \int \limits_x \log \left( \frac{p(x)}{q(x)}  \right) p(x) 
    + \int\limits_{x, y} \log \left( \frac{p(y \giv x)}{q(y \giv x)} \right) p(y \giv x) p(x) \notag \\
  & {} = \mathrm{KL}[p(x) \giv q(x)] + \int_x \mathrm{KL}[p(y \giv x) \giv q(y \giv x)] p(x). 
  \label{eq:new_objective} 
\end{align}

The joint distribution $q(x, y)$ is assumed to be Gaussian in the GF. This
implies that $q(x)$ is Gaussian, and that $q(y \giv x)$ is Gaussian as well with
the mean having a linear dependence on $x$. The parameters of $q(x)$ and $q(y
\giv x)$ are independent, since any two distributions will form a valid joint
distribution $q(x, y)= q(y \giv x) q(x)$. Therefore, the two terms in
\eqref{eq:new_objective} can be optimized independently.


Since the predicted belief $p(x)$ is Gaussian, we can fit it perfectly
$q(x) = p(x)$, resulting in $\mathrm{KL}[p(x) \giv q(x)] = 0$.

The right-most term is the expected KL-divergence between the exact observation
model $p(y \giv x)$ and the approximation $q(y \giv x)$. This makes intuitive
sense: the observation model only needs to be approximated accurately in regions
of the state space which are likely to be visited. Please note that so 
far we have merely taken a different perspective on the standard GF, no
changes have been made yet in this section.

\subsection{Factorization Assumption}

Interestingly, the fact that the exact observation model
\eqref{eq:observation_model} factorizes in the measurement does not imply that
the optimal approximation $q(y \giv x)$ will do so as well. The only change we
apply to the standard GF in this section, is to impose the factorization of the
exact distribution on the approximate distribution $q(y \giv x) = \prod_i q(y^i
\giv x)$. 

Inserting the factorized distributions into \eqref{eq:new_objective}, and making
use of $\mathrm{KL}[p(x) \giv q(x)] = 0$, we obtain
\begin{equation}
  \boxed{
    \begin{aligned}
      \mathrm{KL}[p(x, y) \giv q(x, y)] &= \\
        \sum_i \int\limits_x \mathrm{KL}[p(y^i & \giv x) \giv q(y^i \giv x)] p(x).
    \end{aligned}
  }\label{eq:objizzle}
\end{equation}
The objective function is now a sum over the expected KL-divergences between the
approximate and exact observation models for each sensor. Each of these summands
can be optimized independently. The computational complexity is
reduced and parallelization becomes possible.

\subsection{Finding the Approximate Observation Models}

The joint distribution $q(x,y)$ being Gaussian implies that
the conditionals have the form
\begin{align}
  q(y^i \giv x) = \mathcal{N}(y^i \giv l^i + L^i x, P^i).
  \label{eq:approximate_observation_model}
\end{align}
To find the parameters of $q(y^i \giv x)$ we need to minimize the corresponding 
summand $\int_x \mathrm{KL}[p(y^i \giv x) \giv q(y^i \giv x)] p(x)$ 
of \eqref{eq:objizzle}.
The optimal parameters are retrieved similarly as for the objective \eqref{eq:objective}, 
see \cite{new_perspective} for more details. 

\begin{equation}
  \boxed{
    \begin{aligned}
      l^i &= \mu_{y^i} - \Sigma_{x y^i}^T \Sigma_{xx}^{-1} \mu_x \\
      L^i &= \Sigma_{x y^i}^T \Sigma_{xx}^{-1} \\ 
      P^i &= \Sigma_{y^i y^i} - \Sigma_{x y^i}^T \Sigma_{xx}^{-1} \Sigma_{x y^i}
    \end{aligned}
  }\label{eq:paramizzles}
\end{equation}
The means and covariances are as defined in \eqref{eq:mux}, 
\eqref{eq:sigmax}, \eqref{eq:muy}, \eqref{eq:sigmay} and \eqref{eq:sigmaxy}.

\subsection{Finding the Approximate Posterior}

We have approximated the observation models for each sensor by a linear Gaussian
distribution \eqref{eq:approximate_observation_model}. No further approximations
are required: finding the approximate posterior merely requires some standard
manipulations of Gaussian distributions. The complete observation model is the
product of all the individual observation models
\begin{align}
  & q(y \giv x) = \prod_i q(y^i \giv x) = \\
  & \;\; \mathcal{N}(
    \underbrace{\begin{pmatrix} y^1 \\ \vdots \\ y^M \end{pmatrix}}_{y} \giv
    \underbrace{\begin{pmatrix} l^1 \\ \vdots \\ l^M \end{pmatrix}}_{l} +
    \underbrace{\begin{pmatrix} L^1 \\ \vdots \\ L^M \end{pmatrix}}_{L} x,
    \underbrace{\begin{pmatrix} P^1 & 0 & 0 \\ 0 & \ddots & 0 \\ 0 & 0 & P^{M} \end{pmatrix}}_{P}
    ). \nonumber \label{eq:joint-paramizzles}
\end{align}
Having $q(x)$ and $q(y \giv x)$ we can now apply Bayes rule to express $q(x \giv y)$. 
This operation is simple for Gaussian distributions, and we obtain
\begin{align}
  q(x \giv y) = \mathcal{N}(x \giv \mu_{x|y}, \Sigma_{xx|y})
\end{align}
with the parameters
\begin{equation}
  \boxed{
    \hspace{-0.1cm}
    \begin{aligned}
      \Sigma_{xx|y}^{-1} \! &= \Sigma_{xx}^{-1} + \sum_i (L^i)^T (P^i)^{-1} L^i \\ 
      \mu_{x|y} &= \mu_x \!+\! \Sigma_{xx|y} \sum_i (L^i)^T (P^i)^{-1} (y^i \!-\! l^i \!+\! L^i \mu_x).
   \end{aligned}
   }\label{eq:posterior_parameters}
\end{equation}
Note that for both the precision matrix and the mean of the posterior,
the contributions of each sensor are simply summed up. Unlike for the sequential
GF, these contributions can be computed independently for each sensor, which allows
for parallelization.

\section{Algorithm}\label{sec:algorithm}
The prediction step of the proposed method is identical to the one in 
the standard GF and will therefore not be discussed. A summary of the update step
is presented in Algorithm~\ref{alg:rgf}. The key advantage of the proposed
method is that the $O(N^3)$ computation of the matrix
square root of $\Sigma_{xx}$ has to be performed only once. If we were
to update sequentially in the pixels, we would have to compute 
the square root of the updated covariance matrix each time after incorporating
a pixel. The proposed method, in contrast, generates state samples $\{^k x\}_k$ only
once, outside of the loop over the pixels, see Algorithm~\ref{alg:rgf}.

The remaining computations inside of the loop are of
computational complexity $O(N^2)$. This leads to an improved
computational complexity $O(M N^2 + N^3)$ over the
sequential version $O(MN^3)$.

Furthermore, the computations inside of the for-loop of
Algorithm~\ref{alg:rgf} are independent, allowing
for parallelization, which is not possible with
the sequential GF. While we did not exploit this
possibility in the experiments in this paper, it might become important
as sensors' resolutions and rates keep increasing.

\begin{algorithm}[!tb]
\caption{Proposed Update}
\label{alg:rgf}
\begin{algorithmic}[1]
\Require $\mu_x$, $\Sigma_{xx}$  \Comment{passed from prediction step}
\Ensure  $\mu_{x|y}, \Sigma_{xx|y}$
\State Sample $\sam{x} \sim \mathcal{N}(x \giv \mu_x, \Sigma_{xx}), \; k = 1, \ldots, K.$
\State $D=0, d=0$
\For{each pixel $i = 1, \ldots , M$}
  \State $\sam{y}^i \sim b(y^i \giv \sam{x}), \; k = 1, ..., K.$ \Comment{body of \eqref{eq:observation_model}}
  \State $\mu_{y^i|b} = \mathrm{E}[\{\sam{y}^i\}_k]$ \Comment{empirical mean}
  \State $\Sigma_{y^i y^i|b} = \mathrm{Cov}[\{\sam{y}^i\}_k]$ \Comment{empirical covariance}
  \State $\sam{\varphi}^i = \varphi(\sam{y}^i , \mu_{y^i|b}, \Sigma_{y^i y^i|b}), \; k = 1, ..., K.$ \Comment{\eqref{eq:optimal_feature}}
  \State $\mu_{\varphi^i} = \mathrm{E}[\{\sam{\varphi}^i\}_k]$
  \State $\Sigma_{\varphi^i \varphi^i} = \mathrm{Cov}[\{\sam{\varphi}^i\}_k]$
  \State $\Sigma_{x \varphi^i} = \mathrm{Cov}[\{\sam{x}, \sam{\varphi}^i\}_k] $
  \State $l^i = \mu_{\varphi^i} - \Sigma_{x \varphi^i}^T \Sigma_{xx}^{-1} \mu_x $ \Comment{\eqref{eq:paramizzles}}
  \State $L^i = \Sigma_{x \varphi^i}^T \Sigma_{xx}^{-1}$ \Comment{\eqref{eq:paramizzles}}
  \State $P^i = \Sigma_{\varphi^i \varphi^i} - \Sigma_{x \varphi^i}^T \Sigma_{xx}^{-1} \Sigma_{x \varphi^i} $  \Comment{\eqref{eq:paramizzles}}
  \State $D = D + (L^i)^T (P^i)^{-1} L^i$
  \State $d = d + (L^i)^T (P^i)^{-1} (y^i - l^i + L^i \mu_x)$
\EndFor{}
\State $\Sigma_{xx|y} = (\Sigma_{xx}^{-1} + D)^{-1}$ \Comment{\eqref{eq:posterior_parameters}}
\State $\mu_{x|y} = \mu_x + \Sigma_{xx|y} d$ \Comment{\eqref{eq:posterior_parameters}}
\end{algorithmic}
\end{algorithm}

\setlength{\fboxsep}{0pt}
\setlength{\fboxrule}{0.4pt}

\section{Experimental Setup}
We compare the proposed method to the PF-based object tracker
from \cite{pot}. Before we present the results, we describe the experimental setup.

\subsection{Dataset}
Due to the lack of datasets for 3D object tracking using depth cameras, we
created a dataset which is publicly available \cite{bot}.  
It consists of 37 recorded
sequences of depth images, acquired by an Asus XTION sensor. Each of these
sequences is about two minutes long and shows one out of six objects being moved
with different velocities. To evaluate the tracking performance at different
velocities, the dataset contains sequences with three velocity categories as
show in Table~\ref{table:velocities}. Additionally, the dataset contains
sequences with and without partial occlusion. The camera-object distance range
is between 0.8m and 1.1m.

\begin{table}[!htb]
\begin{tabular}{l*{3}{c}}
              & Velocity 1 &  Velocity 2 & Velocity 3 \\
\hline       
Translational & 5 cm/s   & 11 cm/s  & 21 cm/s \\
Rotational & 10 deg/s & 25 deg/s & 50 deg/s
\end{tabular}
\footnotesize
\caption{Dataset mean translational and rotational velocities of the three 
         categories. }
\label{table:velocities}
\end{table}

\begin{figure}[!htb] 
 \subfigure[]{\fcolorbox{lightgray}{gray}{\includegraphics[height=0.9in]{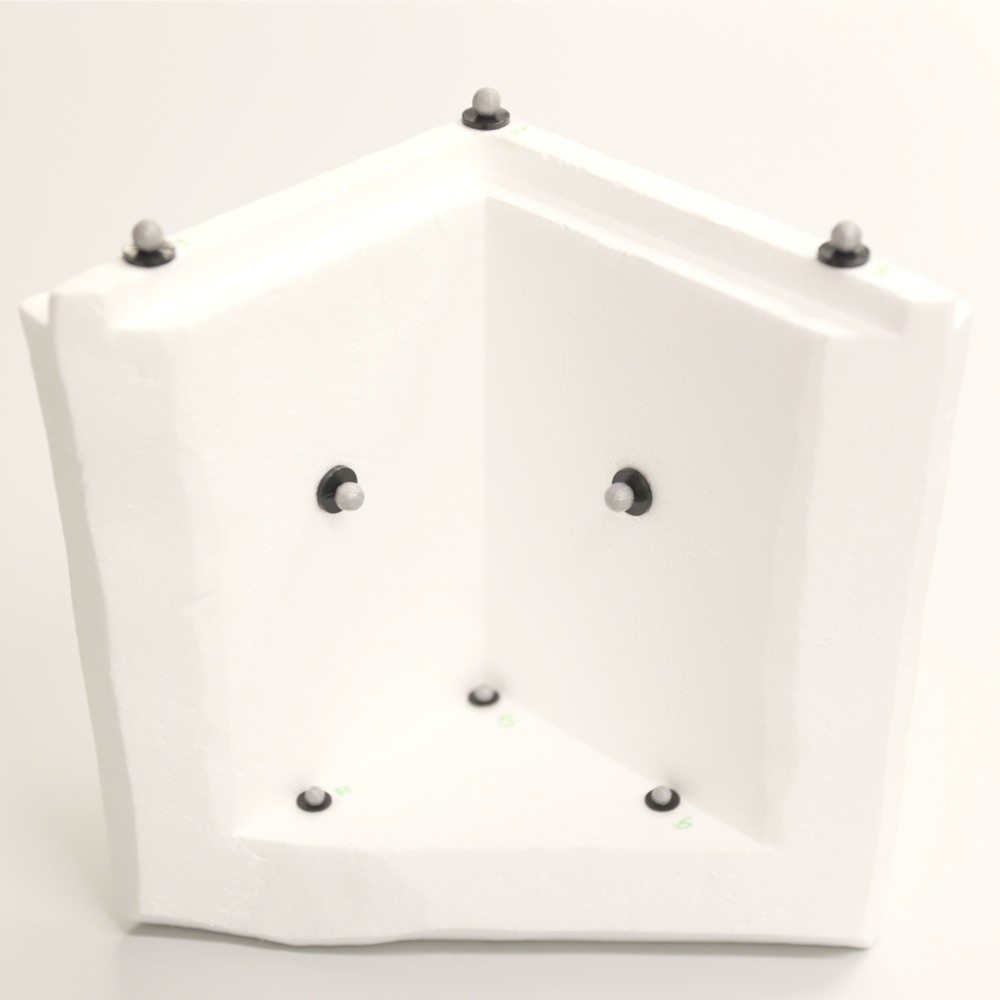}\label{fig:calibration_object}}}
 \subfigure[]{\fcolorbox{lightgray}{gray}{\includegraphics[height=0.9in]{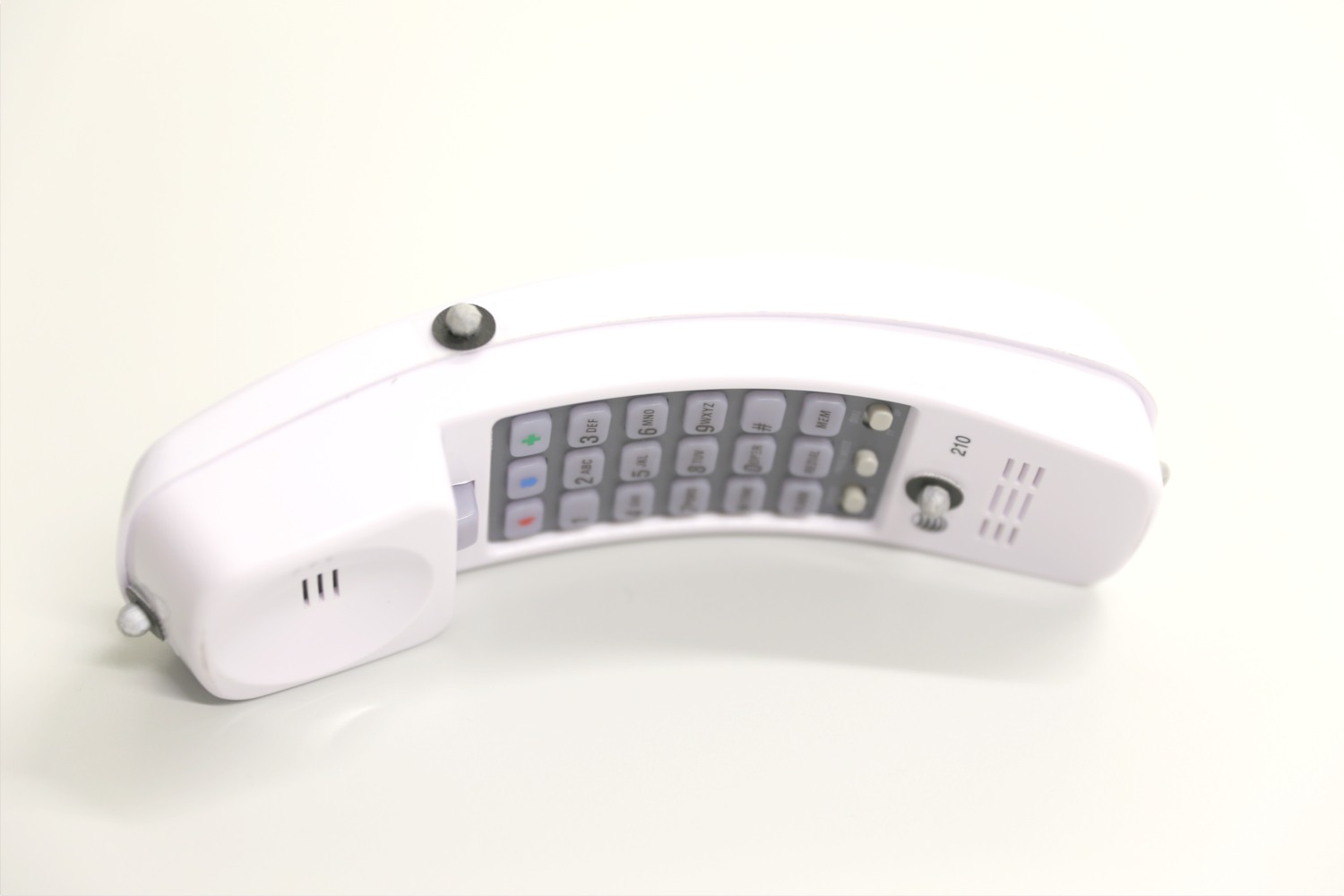}\label{fig:handset}}}
 \subfigure[]{\fcolorbox{lightgray}{gray}{\includegraphics[height=0.9in]{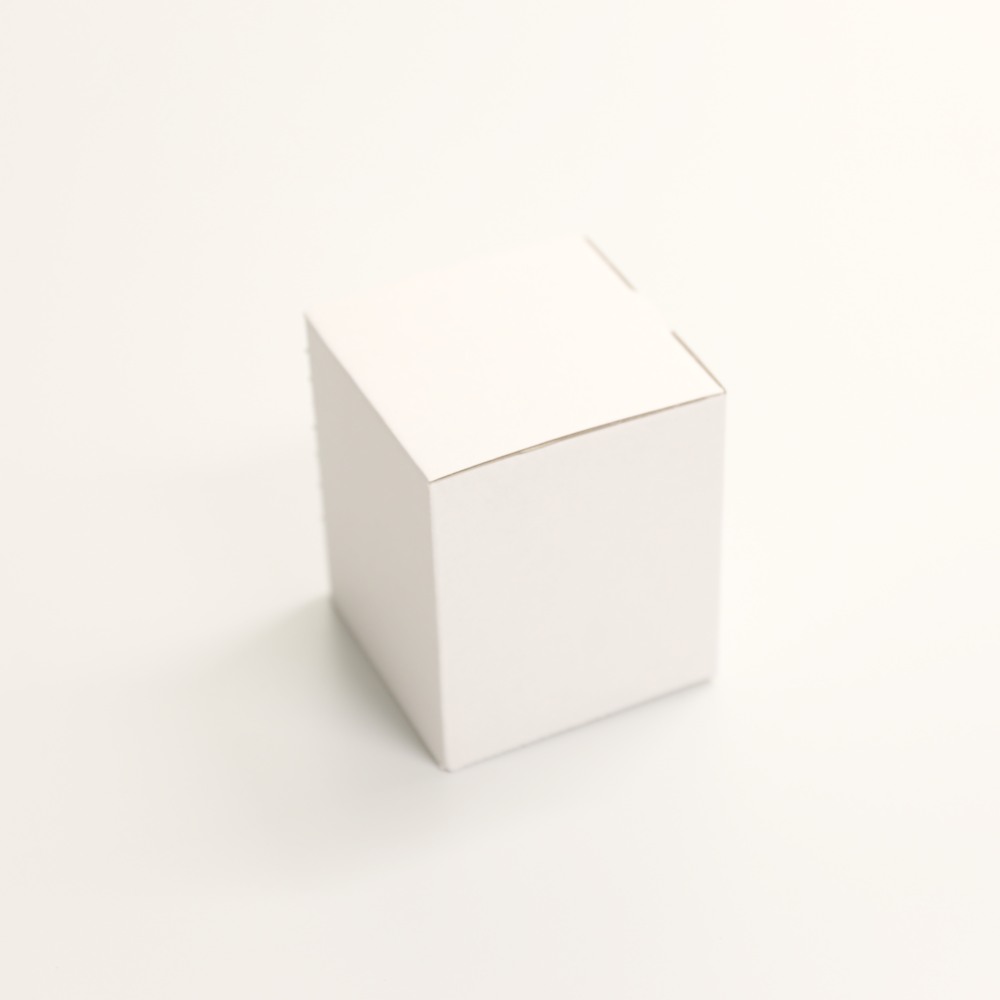}}}
 \newline
 \subfigure[]{\fcolorbox{lightgray}{gray}{\includegraphics[height=0.9in]{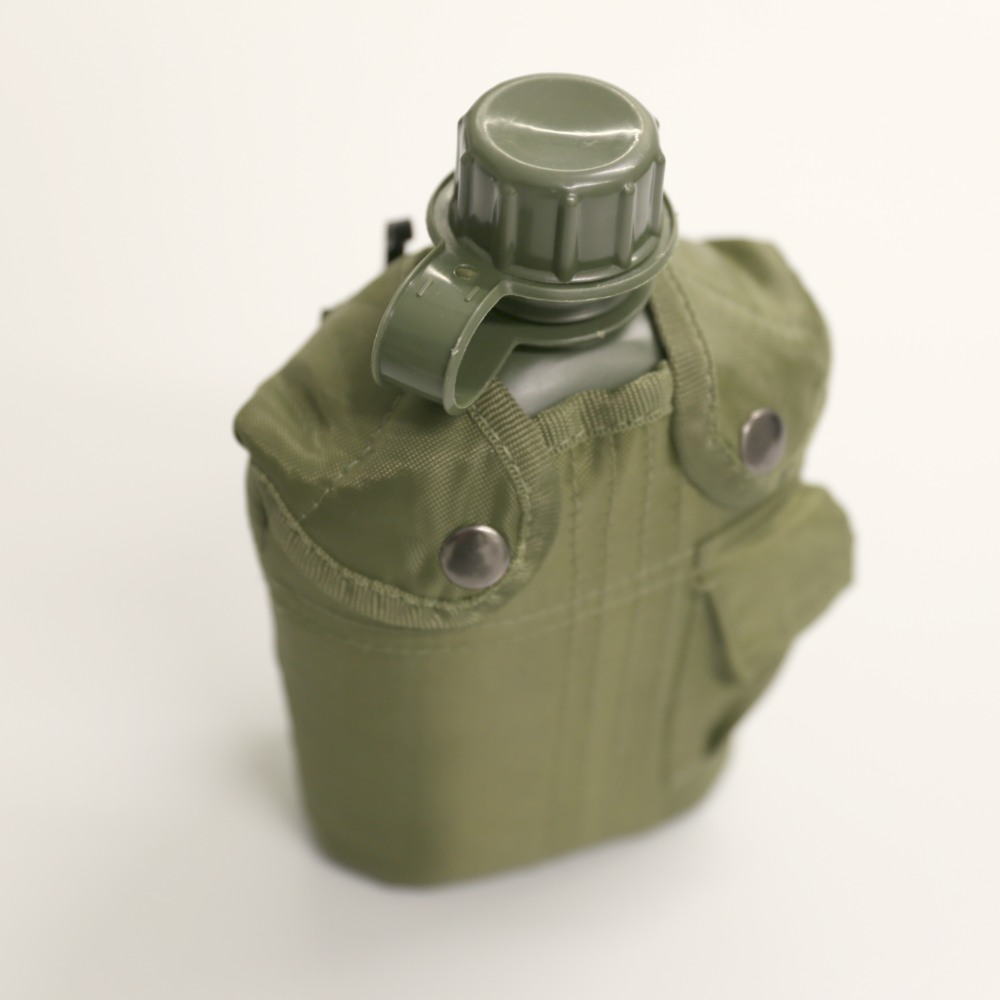}}}
 \subfigure[]{\fcolorbox{lightgray}{gray}{\includegraphics[height=0.9in]{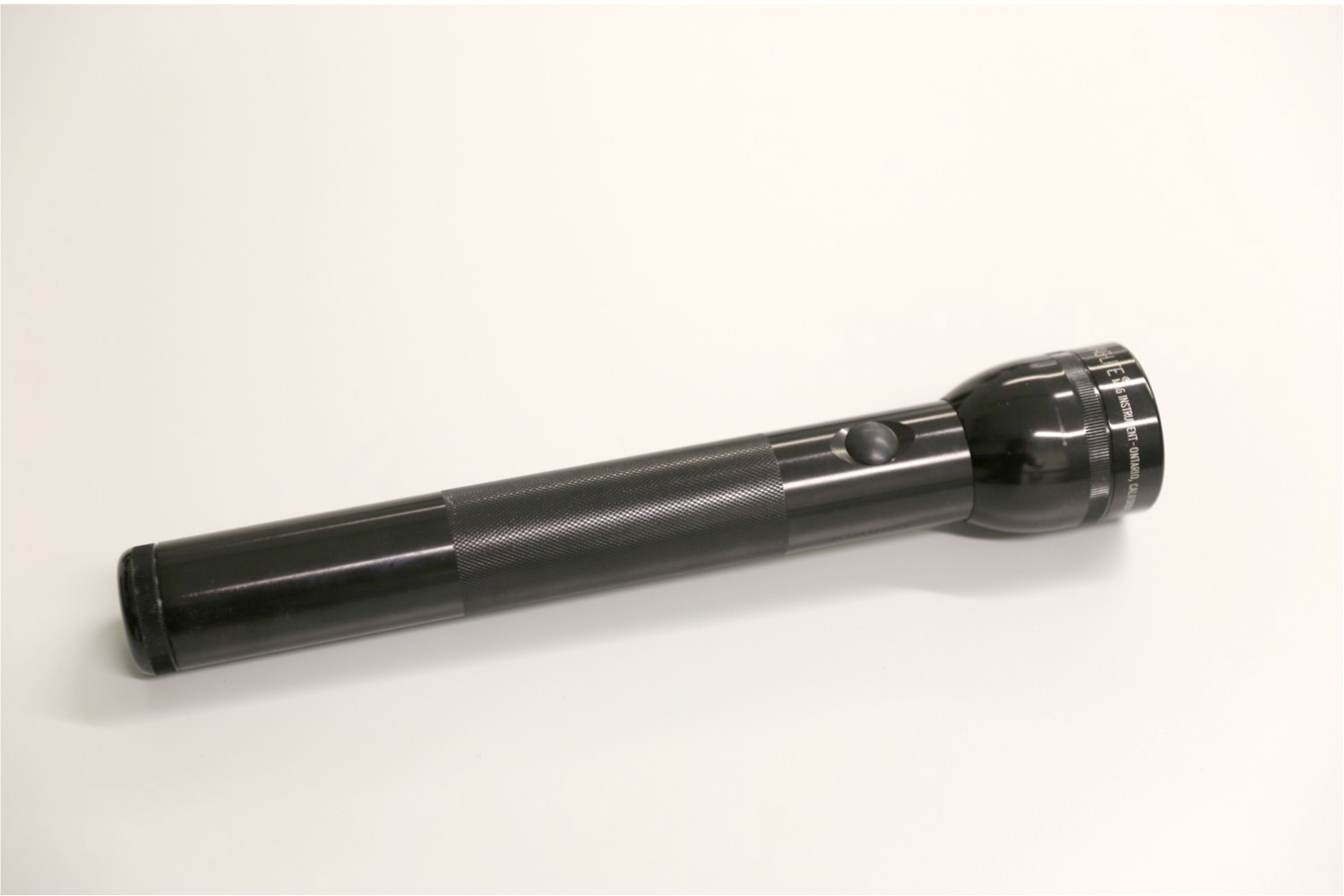}}}
 \subfigure[]{\fcolorbox{lightgray}{gray}{\includegraphics[height=0.9in]{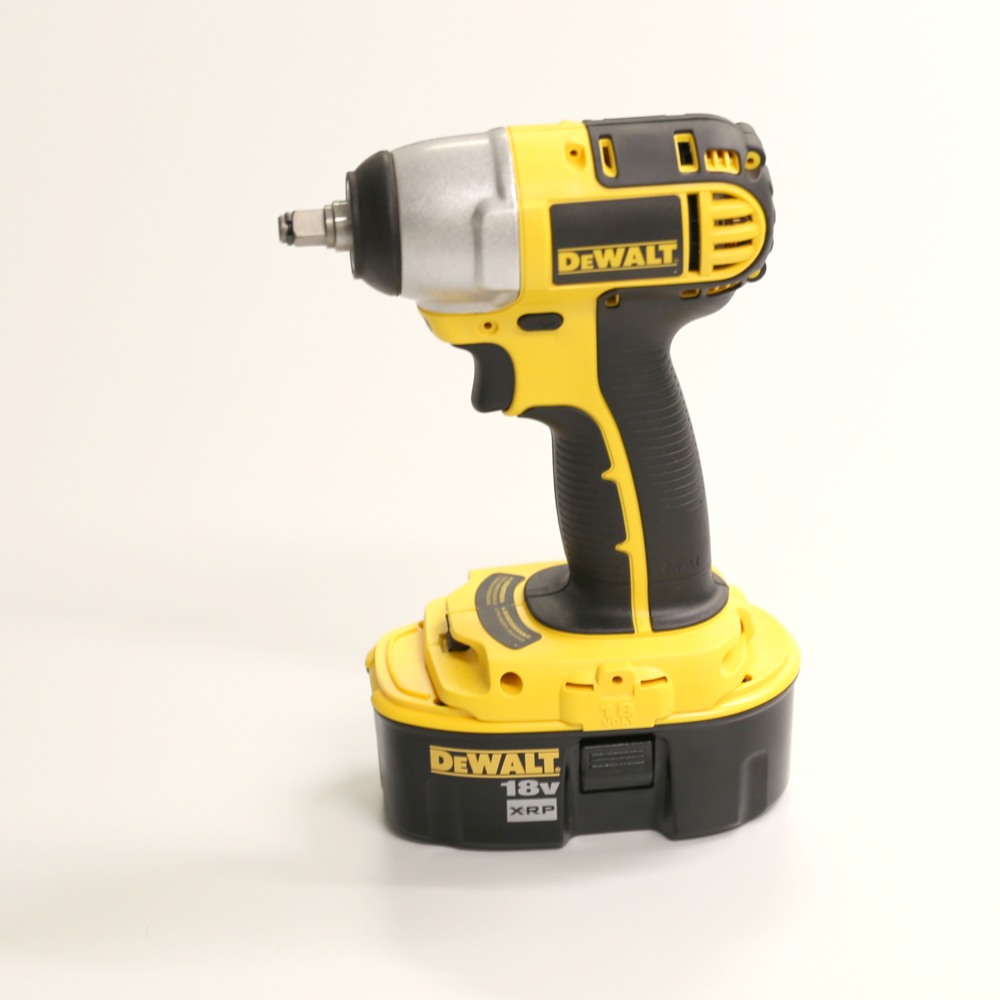}}}
 \caption{Experiment and dataset objects}
 \label{fig:objects}
\end{figure}

For each sequence we also provide a mesh model of the tracked object, 
along with the ground-truth poses for each depth-image frame. The ground-truth has
been acquired by a Vicon system by placing 4mm markers as shown in figures 
\ref{fig:handset} and \ref{fig:calibration_object}.


\subsection{Implementation Details}
%
%
%
%
%


The following parameters were chosen empirically for a processing rate of 30Hz.
The process model has a
$1$~mm standard deviation in translational noise $\epsilon^v_t$ and $0.01$~rad
standard deviation in rotational noise $\epsilon^\omega_t$. The observation
noise was set to $1$~mm standard deviation. The tail probability was set to
$w=0.1$. The RGF is not sensitive to the tail probability as discussed in \cite{rgf}.
%

To achieve real-time performance (30Hz) on a single CPU core, we used a
downsampling factor of 10 for the depth image with a VGA resolution of $640
\times 480$ pixels. This results in a 3072 dimensional observation. 
For numeric integration we use the canonical Unscented Transform with the
parameters $\alpha = 1$, $\beta=2$, and $\kappa=0$. The implementations of both
the proposed method and the PF-based method are available in the open-source
software package {\em Bayesian Object Tracking} \cite{bot}.

%

\section{Experimental Results}


In the following experiments we show the importance of robustifying the GF, an
example of tracking under heavy occlusion, and a quantitative evaluation of the
tracking performance on the full dataset of the proposed method compared to a
PF-based tracker \cite{pot}. A summary video illustrating the tracking
performance is available at
\href{https://www.youtube.com/watch?v=mh9EIz-Tdxg}{
https://www.youtube.com/watch?v=mh9EIz-Tdxg}.

\subsection{Importance of the Robustification}\label{sec:comparison_to_standard_gf}

\begin{figure}[tb]
  \centering
  \includegraphics[width=\linewidth]{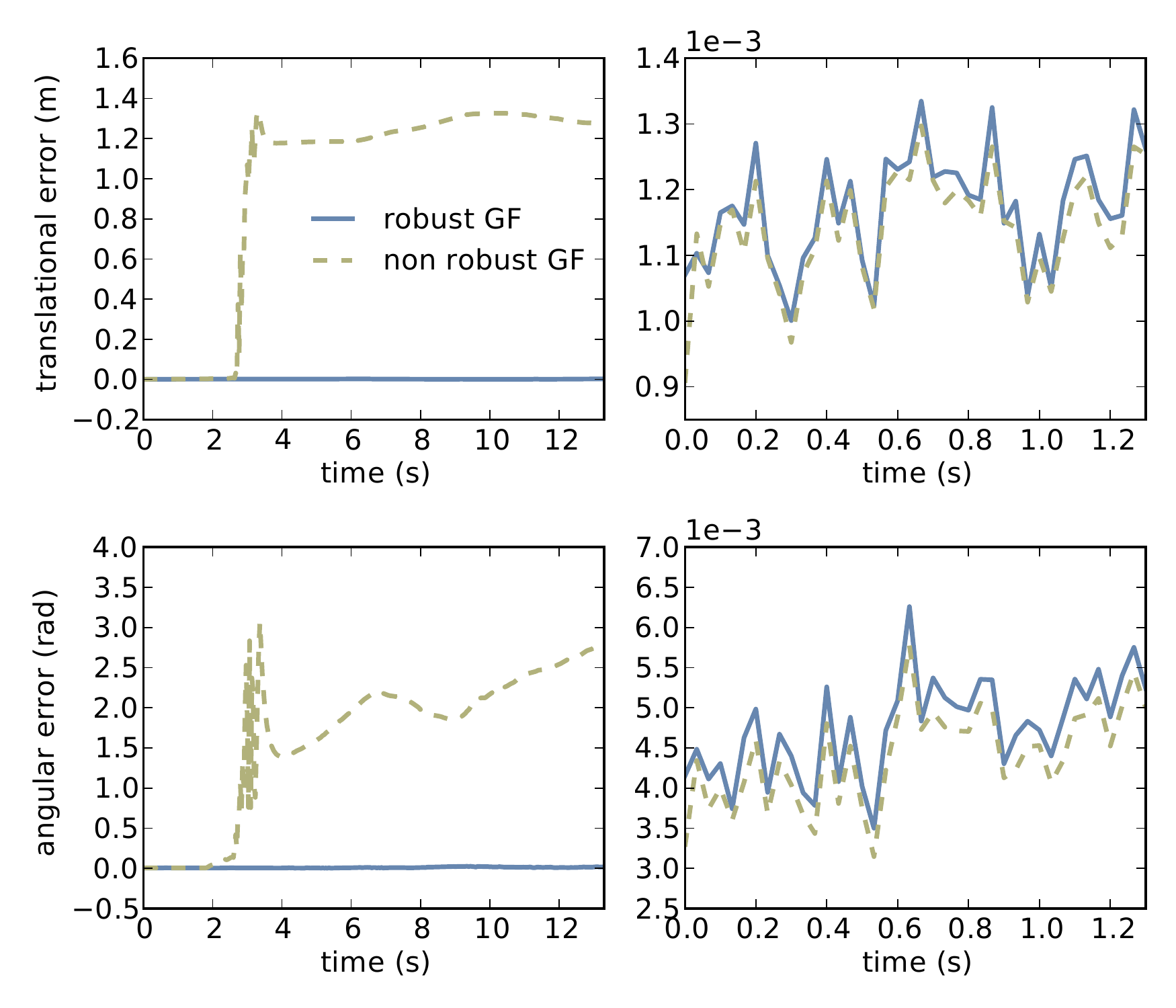}
  \caption{
    The left column shows the error over time of the robust GF and the standard GF. 
    After 2 seconds the object is partially occluded. The right column shows a zoom 
    on the beginning of the time series, before the occlusion occurs.
    \label{fig:RGF-GF-time}}
\end{figure}

%
%
 
We anticipated in Sections \ref{sec:model} and \ref{sec:fat_tails} that the
standard GF does not work with heavy-tailed measurements.  
Instead, the RGF can handle this kind of measurement noise, and therefore allows
to take a GF-based approach to filtering using depth data. To illustrate this
empirically, we compare the proposed method to a standard GF. The observation
model in the proposed method is given by \eqref{eq:observation_model}. For the
standard GF we choose the same observation model without its tail. In practice
this is achieved by executing exactly the same method, with a tail weight $w=0$.

We use as input data a recording of the impact wrench (Figure~\ref{fig:impact})
standing still on the table. After a couple of seconds, a partial occlusion
occurs.

We run both filters and calculate translational error and angular error between
the estimated poses and the ground-truth poses. In Figure~\ref{fig:RGF-GF-time}
we show these errors over time.

Before the occlusion occurs, both filters have very similar performance. Then,
in the presence of outliers, the standard GF reacts strongly and its error
increases abruptly, while the proposed method remains steady and its error low for
the whole sequence.



\subsection{Filtering under Heavy Occlusion}

\begin{figure}[tb]
  \centering
  \includegraphics[width=\linewidth]{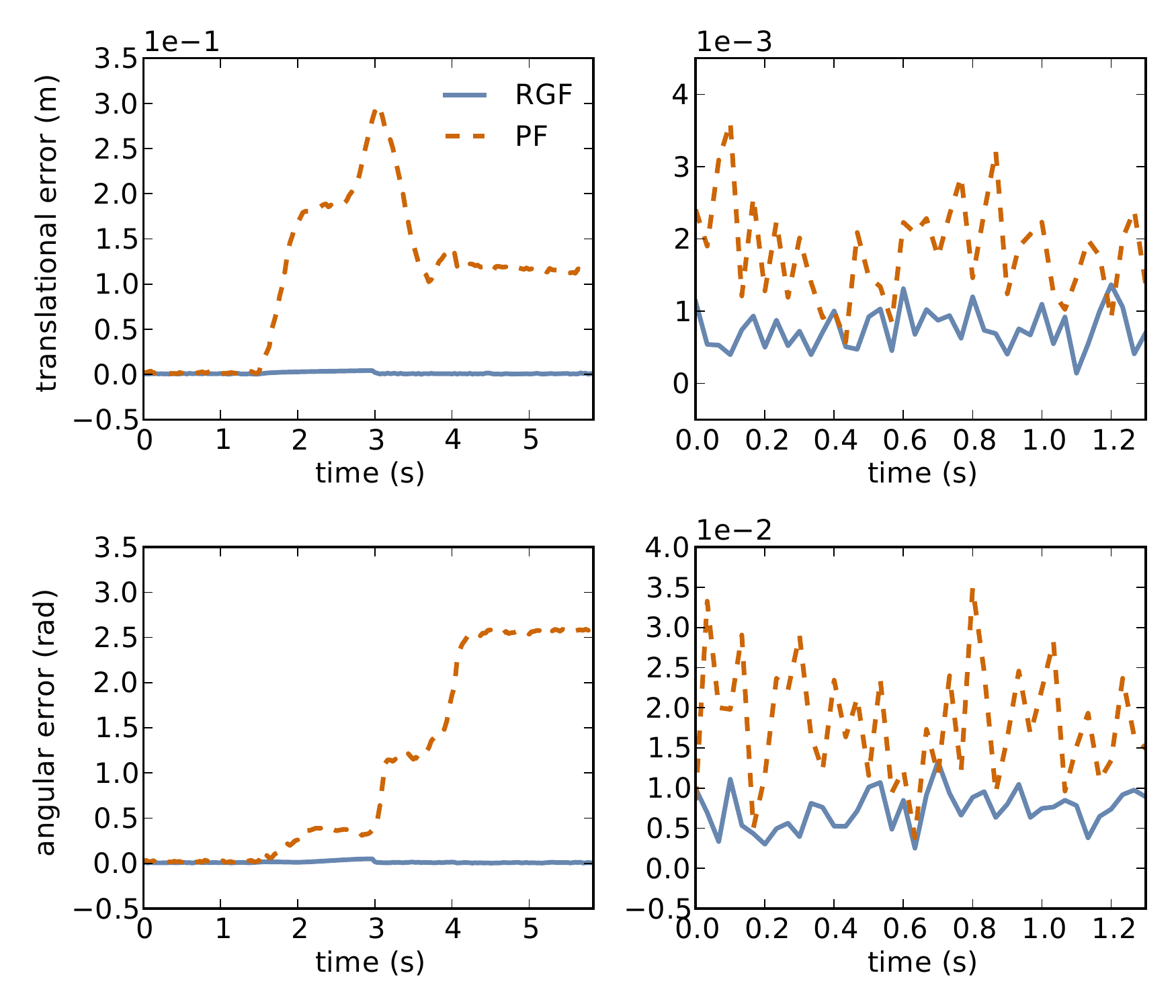}
  \caption{
    Error over time. 
    Robust GF vs. PF. 
    On the right column, a zoom of the first time steps. 
    In the case of a full occlusion, the PF tends to drift, while the robust GF is more stable. 
    \label{fig:RGF-PF-time}}
\end{figure}

Here compare the proposed GF-based method to the PF-based tracker proposed in
\cite{pot}. We run both on a sequence showing an impact wrench standing on a
table, while being partially occluded. After a couple of seconds, there is a
total occlusion lasting several seconds. 

Again, we calculate translational and
angular error with regards to ground-truth
annotations. Figure~\ref{fig:RGF-PF-time} shows tracking error over time.

Both approaches perform well in the presence of partial occlusion, in the
beginning of the sequence. However, the zoom on the right-hand side of the
figure reveals a key advantage of GF-based methods over PF-based methods: The
RGF is much more accurate and more steady. This is particularly important when
this estimate is to be used for controlling a robot.

When the object becomes entirely occluded, the PF starts drifting randomly due
to its stochastic nature. The RGF however reacts in a more deterministic manner:
When only outliers are received, these are soft-rejected and the estimate simply
reverts to the prediction \cite{rgf}, and remains therefore steady.

%

%
%

\subsection{Quantitative Comparison}

For this experiment, we run the proposed method and the PF on the 37 sequences
of the dataset. In the case of the PF, we aggregate the results of 30
independent runs. This is not necessary for the robust GF since the
implementation is based on the UT and is therefore deterministic.

Both trackers are reset to the ground-truth after every 5 seconds to avoid
comparisons between the two while one has lost track.

\begin{figure}[tb]
  \centering
  \includegraphics[width=\linewidth]{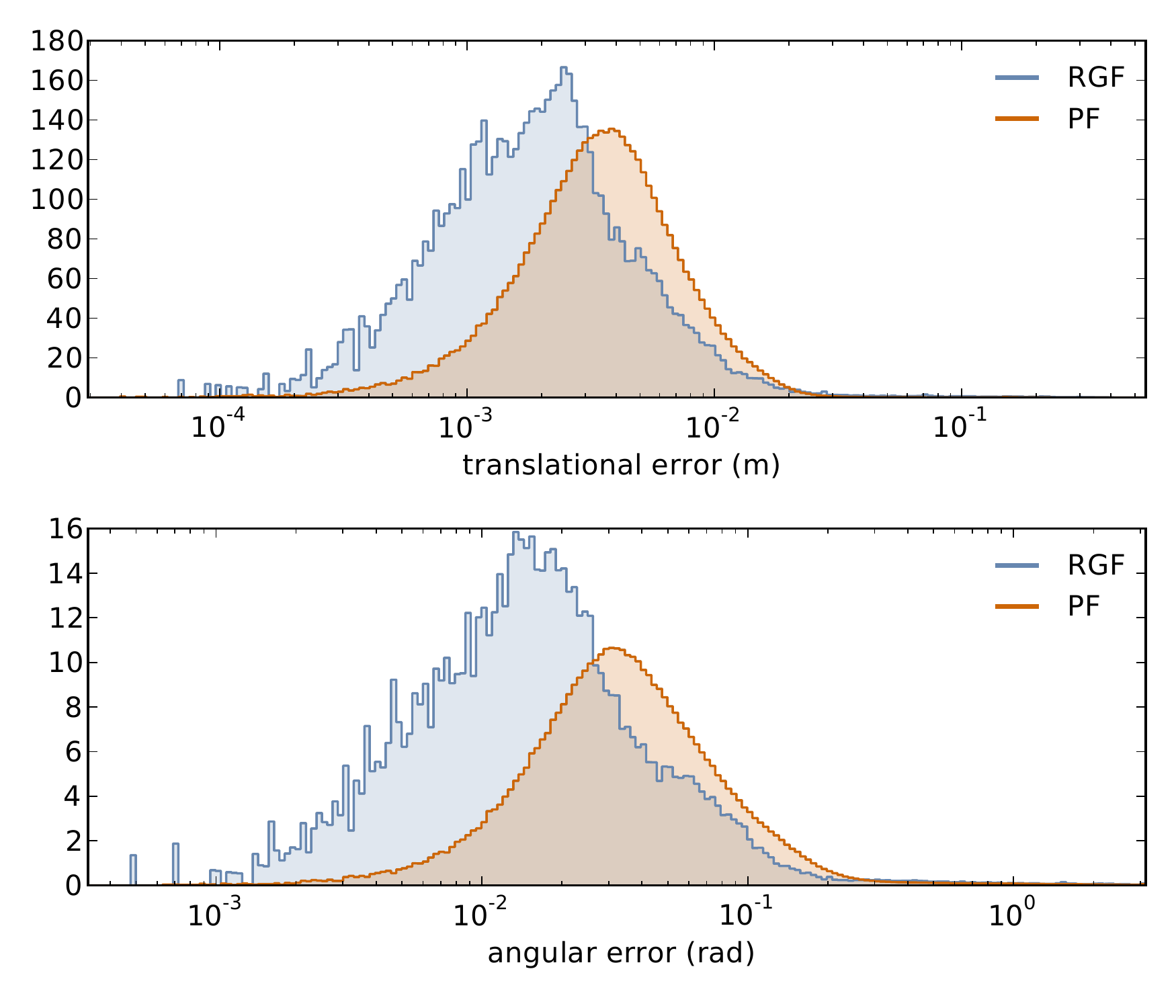}
  \caption{
    Distribution of translational error (top) and angular error (bottom)
    obtained with the robust GF and the PF (aggregation over 30 runs). The robust
    GF is more precise than the PF for small errors.
    \label{fig:RGF-PF-30}}
\end{figure}

At each time step, we evaluate the translational and the angular error as in
previous experiments. The results are summarized in Figure~\ref{fig:RGF-PF-30},
which shows the distribution of the errors for both methods. We can see that the
RGF yields more accurate estimates. However, the proposed method looses track a
little more often than the PF-based method. This only happens very rarely and is
therefore not visible in the density in Figure~\ref{fig:RGF-PF-30}.
Nevertheless, these rare large errors lead to a higher mean error, see
Table~\ref{table:errortable}. However, the median error of the proposed method
is smaller than the one of the PF-based method.

Summarizing, we can say that the PF-based method \cite{pot} is a little more
robust, but that the proposed GF-based method is more accurate and yields
smoother estimates. The properties of the two filters are complementary. For
tracking of fast motion the PF-based method might be preferable, but for
accurate visual servoing the proposed method is more suitable.

\begin{table}[tb]
\begin{tabular}{l*{3}{c}}
                           & RGF & PF  \\
\hline       
Translational error  mean [mm]  & 0.03008 & 0.00956 \\
Translational error  median [mm] &  0.00489 & 0.00556 \\
Angular error mean [deg]      & 0.39644 &  0.29044 \\
Angular error median [deg]    & 0.06076 & 0.07146 \\
\end{tabular}
\footnotesize
\caption{Mean and median errors for RGF and PF.}
\label{table:errortable}
\end{table}

\section{Conclusion}
In this paper we apply a Gaussian Filter to the problem of 3D object tracking
from depth images. There are two major obstacles to applying the GF to this
problem: First of all, the GF is inherently non-robust to outliers, which are
common in depth data. We address this problem by applying the robustification
method proposed in \cite{rgf}.

Secondly, the complexity of the standard GF is prohibitive for the 
high-dimensional measurements obtained from a depth camera. We propose 
a novel update which has a complexity of $O(M N^2 + N^3)$, where
$M$ is the number of sensors, and $N$ is the dimension of the state space.
Furthermore, the proposed update is eligible for parallelization since
the information coming from each pixel is processed independently.

The experimental results illustrate the advantage of GF based methods
over PF-based methods. They provide often smoother and more accurate estimates.
These are very important properties if the estimates are used to control 
a robot.

\scriptsize{
  \bibliographystyle{plainnat}
\bibliography{bibliography}
}

\end{document}